\documentclass[letterpaper,10pt,conference]{ieeeconf}
\IEEEoverridecommandlockouts
\usepackage{graphicx}
\usepackage{amsmath,amssymb}
\usepackage{booktabs,multirow}
\usepackage{subcaption}
\usepackage{microtype}
\usepackage{cite}
\usepackage[hidelinks]{hyperref}
\graphicspath{{./}{figures/}}

\newcommand{\iu}{\mathrm{j}}
\newcommand{\ee}{\mathrm{e}}
\newcommand{\T}{\mathsf T}
\newcommand{\HH}{\mathsf H}
\DeclareMathOperator{\diag}{diag}
\DeclareMathOperator{\tr}{tr}
\DeclareMathOperator{\vect}{vec}
\DeclareMathOperator*{\argmin}{arg\,min}

\title{\LARGE\bf   Single-Radar Bilateral 3D Near-Field Imaging for Mobile Robots}
\newif\ifshowaffil
\showaffilfalse
\author{Yingqi Wang$^{*}$, Tingwei Chen$^{*}$, Zhongqin Wang, Min Xu%
\thanks{$^{*}$These authors contributed equally to this work.}%
\ifshowaffil
  \thanks{All authors are with AFFILIATION.
          {\tt\small EMAIL1, EMAIL2, EMAIL3, EMAIL4}}%
\fi
}
\begin{document}
\maketitle
\thispagestyle{empty}
\pagestyle{empty}

\begin{abstract}
Mobile robots operating in homes need reliable 3D perception of nearby furniture and surrounding structures, yet recovering this information on both sides of the robot's path with a single forward-mounted millimeter-wave radar remains challenging because of near-field wavefront curvature, viewpoint-dependent target visibility, and limited elevation diversity. This paper presents a motion-assisted framework for reconstructing structures on both sides of a robot's trajectory. Our near-field model relates scene reflectivity to stepped-frequency multiple-input multiple-output (MIMO) observations across frequencies and robot positions using spherical paths from transmitter (Tx) to scatterer to receiver (Rx). Bilateral Tx combining provides wideband directional weighting, while sliding-aperture back-projection coherently focuses the measurements using near-field propagation distances. Top-$K$ power fusion combines local images with different visibility. The reconstruction provides lateral--forward support for multi-layer height extraction, followed by guarded adaptive height refinement using the original MIMO channels. Experiments in controlled, office/study, and corridor environments produce mean lateral--forward localization errors of $4.65$, $5.63$, and $5.98$~cm, respectively, with peak-to-background ratios (PBRs) exceeding $13$~dB. The framework recovers bilateral 3D scattering structure from one forward pass using a single radar, without relying on learned representations of the scene.
\end{abstract}

\section{Introduction}
Mobile robots navigating homes need 3D information about nearby furniture, household objects, and walls, including those on both sides of a path through a room or corridor. Millimeter-wave (mmWave) radar offers compact sensing that measures range and motion without relying on ambient illumination~\cite{harlow2024survey,prabhakara2023radarhd}. However, compact multiple-input multiple-output (MIMO) arrays have limited angular resolution, especially in elevation, and indoor reflections and multipath complicate the interpretation of radar echoes. Processing each frame independently often yields sparse, fragmented detections, while thresholding can discard weaker echoes associated with nearby objects~\cite{huang2025foundational}. The spatial extent and vertical structure of the surrounding scene therefore remain difficult to recover from individual radar frames.

Existing approaches integrate radar imaging with odometry or probabilistic mapping~\cite{ritterbusch2024synthetic,huang2026probabilistic}, exploit distributed cross-device channels~\cite{sun2026muldar}, or learn richer representations with LiDAR or visual--inertial supervision~\cite{prabhakara2023radarhd,luan2024radardiffusion,fan2024mmemp}. Distributed sensing and learned reconstruction involve additional hardware, calibration, or training requirements. For static-scene reconstruction, robot motion provides observations from different positions that can be combined through synthetic-aperture processing. 
At short indoor distances, coherent integration is sensitive to wavefront curvature across both the antenna and motion apertures. Spherical-wave modeling retains the transmitter–scatterer–receiver geometry needed to align complex echoes across robot positions~\cite{smith2022irregular,miran2021sparse}, whereas propagation-model errors lead to spatial defocusing.
Small propagation-distance errors can misalign these echoes, spreading a compact scattering response over neighboring locations instead of reinforcing its true position. This motivates bilateral 3D reconstruction from one forward-mounted radar using robot motion and explicit near-field propagation models for coherent spatial focusing.

For this single-radar configuration, an unmanned ground vehicle (UGV) moving through a static indoor scene provides different viewing angles along its forward trajectory. A lateral object may produce strong echoes over only part of the path, so its response can vary substantially between local reconstructions formed from different trajectory segments. Averaging all local reconstruction powers can then weaken otherwise distinct object responses. Objects near the direction of travel also exhibit less angular variation and can remain difficult to localize. Meanwhile, a straight, level trajectory extends the aperture mainly horizontally, leaving height estimation largely dependent on the physical antenna array. Furniture and nearby household objects can also produce multiple vertically separated echoes within the same horizontal region. These properties motivate selecting informative local reconstruction responses and using the recovered horizontal locations to guide the extraction of multiple scattering heights.

In this work, we present BilatSAR, a motion-assisted framework for bilateral 3D indoor reconstruction using a UGV with a single forward-mounted millimeter-wave radar. A spherical-wave model relates scene reflectivity to measurements across transmit (Tx) and receive (Rx) antennas, frequencies, and robot positions. Frequency-dependent weights digitally combine the recorded Tx channels toward the left and right sides of the path. This angular weighting does not select a focal distance; subsequent near-field back-projection uses candidate-dependent propagation distances to align echoes over short, overlapping trajectory segments. Top-$K$ fusion combines the strongest local reconstruction powers at each scene location. The resulting volume identifies horizontal support for extracting multiple scattering heights, allowing vertically separated scatterers to share a horizontal location. Near-field minimum-variance distortionless-response (MVDR) processing then refines these heights using the original MIMO channels while keeping horizontal positions fixed. Local response checks guard each update and retain the initial height when refinement is unreliable. These stages address visibility changes and limited vertical aperture.

This paper makes the following contributions:
\begin{itemize}
\item We formulate a near-field MIMO measurement model linking scene reflectivity to observations across antennas, frequencies, and robot positions. The formulation distinguishes the effective-transmitter approximation used in back-projection from the full transmitter--scatterer--receiver paths used for height refinement.
\item We develop a bilateral reconstruction method integrating bilateral Tx combining, local near-field focusing, and visibility-dependent power fusion, followed by multi-layer height extraction at detected horizontal locations. Guarded near-field MVDR then refines candidate heights using the original MIMO channels while keeping horizontal positions fixed.
\item We demonstrate single-radar bilateral 3D sensing in controlled
household-object, office/study, and corridor scenes, achieving mean
lateral--forward errors below 6~cm and PBR above 13~dB. Across all three scenes, bilateral Tx combining improves background separation, while mean localization errors remain below those of frame-wise MIMO
Range--AoA.
\end{itemize}

\section{Related Work}
We review near-field imaging, motion-assisted perception, and volumetric reconstruction for 3D radar sensing.

\subsection{Near-Field and High-Resolution Radar Imaging}
Physical, virtual, and synthetic apertures improve on frame-wise mmWave angle estimation. MIMO-SAR supports irregular scans~\cite{smith2022irregular} and efficient near-range handheld imaging via factorized back-projection~\cite{wang2025handheld}. Automotive SAR examines viewing geometry, angular ambiguity, and motion accuracy~\cite{tebaldini2022urban}. RImID uses motion-compensated inverse synthetic aperture radar (ISAR) imaging for user reidentification~\cite{wang2025rimid}. Spherical-wave focusing accounts for phase curvature, while resolution depends on aperture extent and viewing diversity. Accurate focusing cannot recover viewing angles absent from the acquired measurements. Our forward-mounted sensor requires bilateral processing despite its limited vertical aperture.

\subsection{Motion-Assisted Robot Perception}
Robotic SAR systems integrate radar-inertial odometry with image formation~\cite{ritterbusch2024synthetic}, quantify usable coherent apertures on wheeled robots~\cite{ritterbusch2024indoor}, and couple localization and mapping with SAR processing~\cite{grebner2024simultaneous}. Phase-autocorrelation correction compensates motion-induced phase errors in near-field SAR back-projection~\cite{cheng2026phase}. Our fusion targets changing scatterer visibility, not trajectory errors. Other systems combine SAR with probabilistic mapping~\cite{huang2026probabilistic} or use distributed cross-device channels~\cite{sun2026muldar}. Forward-looking sensing has limited angular diversity along the travel direction; MIMO processing and Doppler beam sharpening address left--right ambiguity and robust 3D estimation~\cite{yuan2022novel,yuan2024robust}. We target bilateral sensing from one forward pass, combining local spherical-wave focusing and visibility-aware fusion to establish horizontal support for subsequent height estimation.

\subsection{Volumetric Reconstruction and Scene Understanding}
Near-field volumetric reconstruction can be formulated as a linear inverse problem. Fast-multipole sparse reconstruction evaluates the sensing operator and adjoint without dense storage~\cite{miran2021sparse}, while physics-based learned methods combine back-projection with a 3D reconstruction network~\cite{manisali2024physics}. RadarHD learns high-resolution representations from LiDAR-paired radar data~\cite{prabhakara2023radarhd}, whereas Radar-diffusion uses LiDAR-supervised diffusion to suppress ghosts and densify 3D radar point clouds~\cite{luan2024radardiffusion}. mmEMP uses reconstructed dynamic visual features for supervision and temporal spatial consistency to reject spurious points~\cite{fan2024mmemp}. Prediction, diffusion, and upsampling recover occupancy or completed point clouds~\cite{ding2024radarocc,zhang2024dense,mopidevi2024rmap,huang2025foundational}. Rascene reconstructs 3D scenes from mmWave communication signals through spatially adaptive multi-frame fusion~\cite{song2026rascene}. Neural representations incorporate radar signal formation: DART uses Doppler tomography for novel-view synthesis~\cite{huang2024dart}, and Radar Fields models raw FMCW observations in frequency space~\cite{borts2024radarfields}. Our method uses the measured wavefield and known trajectory to establish spatial support and refine multiple heights, recovering dominant scattering structure without assuming complete surfaces or uniform resolution.

\begin{figure*}[t]
\centering
\includegraphics[width=\textwidth]{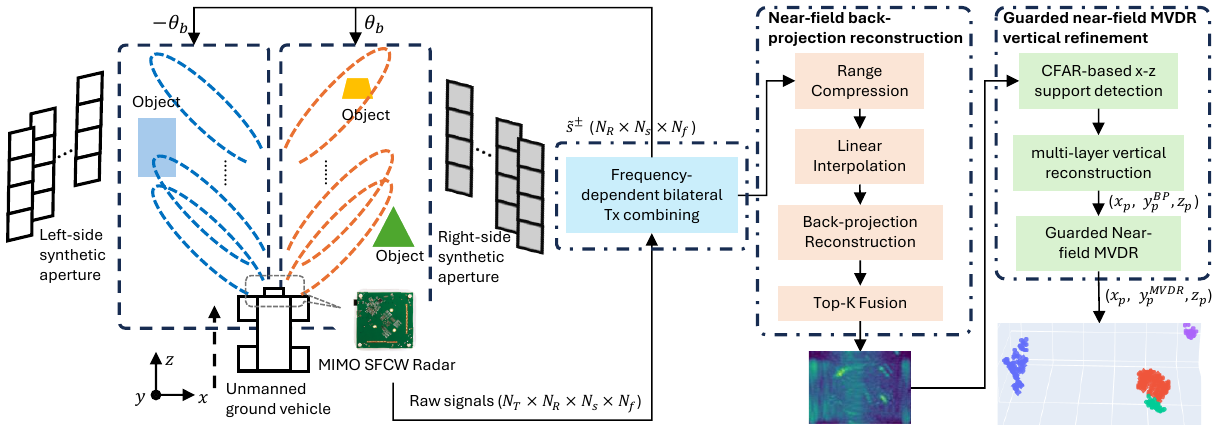}
\caption{BilatSAR overview: frequency-dependent Tx combining, sliding-aperture near-field back-projection, and Top-$K$ fusion establish lateral--forward support, followed by multi-layer height extraction and guarded MVDR refinement.}
\label{fig:diagram}
\vspace{-1.5em}
\end{figure*}

\section{Near-Field Signal Model}
\label{sec:model}
We model near-field MIMO measurements and examine geometric limits on focusing and aperture diversity.

\subsection{Geometry and Assumptions}
\label{sec:geometry}
We consider a forward-mounted stepped-frequency continuous-wave
(SFCW) MIMO radar using $N_T$ active Tx and $N_R$ Rx elements
over frames $n=0,\ldots,N_s-1$. World axes $x$, $y$, and $z$
denote lateral, vertical, and forward directions. The mean Tx phase
center at frame $n$ is $\mathbf p_n=[0,0,z_n]^\T$, where $z_n$
follows the known robot motion, and the radar orientation is fixed
along the approximately straight path. Calibrated Tx and Rx offsets
$\boldsymbol\delta_i^{\mathrm{Tx}}$ and
$\boldsymbol\delta_j^{\mathrm{Rx}}$ are expressed in this world frame
relative to the mean Tx center; their physical 3-D coordinates are
used directly without assuming a uniform array. At
$\mathbf q=[x,y,z]^\T$, the antenna geometry gives
\begin{equation}
\resizebox{\dimexpr\columnwidth-3em\relax}{!}{$\displaystyle
r_{i,n}^{\mathrm{Tx}}(\mathbf q)=\|\mathbf q-\mathbf p_n-\boldsymbol\delta_i^{\mathrm{Tx}}\|_2,\quad
r_{j,n}^{\mathrm{Rx}}(\mathbf q)=\|\mathbf q-\mathbf p_n-\boldsymbol\delta_j^{\mathrm{Rx}}\|_2.
$}
\label{eq:element_ranges}
\end{equation}
The total propagation path from Tx $i$ through $\mathbf q$ to Rx $j$ is
\begin{equation}
R_{i,j,n}(\mathbf q)
=r_{i,n}^{\mathrm{Tx}}(\mathbf q)+r_{j,n}^{\mathrm{Rx}}(\mathbf q).
\label{eq:bistatic_range}
\end{equation}
The propagation delay equals $R_{i,j,n}(\mathbf q)$ divided by wave speed $c$. Each path is evaluated from the candidate and robot positions, without an arrival-angle approximation.

The radar records $N_f$ uniformly spaced frequencies. For indices $m=0,\ldots,N_f-1$, the frequencies are $f_m=f_0+m\Delta f$, with wavenumbers $\kappa_m=2\pi f_m/c$. We assume calibrated complex channels, negligible robot motion during one frequency sweep, and a static scene within each coherent aperture. Reflectivity is approximately constant within each aperture; aspect-dependent visibility and multipath can cause model mismatch. Bold lowercase and uppercase symbols denote vectors and matrices, respectively. Superscripts $(\cdot)^\T$, $(\cdot)^\HH$, and $(\cdot)^*$ denote transpose, conjugate transpose, and complex conjugation. The imaginary unit satisfies $\iu^2=-1$. The operator $\diag(\mathbf v)$ places the entries of $\mathbf v$ on a diagonal, and $\|\cdot\|_2$ denotes the norm of a Euclidean vector.

\subsection{Near-Field MIMO Measurement Model}
\label{sec:matrixmodel}
Discretize the imaging region into $N_v$ cells centered at $\mathbf q_g$, with integrated complex reflectivities $\rho_g$ collected in $\boldsymbol\rho$. The Tx--Rx measurement at frame $n$ and frequency $f_m$ is
\begin{equation}
s_{i,j,n}(f_m)=\sum_{g=1}^{N_v}\rho_g
\ee^{-\iu\kappa_mR_{i,j,n}(\mathbf q_g)}+\eta_{i,j,n,m}.
\label{eq:measurement_model}
\end{equation}
The residual $\eta_{i,j,n,m}$ accounts for noise and deviations from the phase-only, single-scattering assumption. This model describes geometric phase for spatial focusing; absolute reflectivity is not estimated in this pipeline.

At each frame and frequency, the $N_R\times N_T$ measurement matrix $\mathbf S_{n,m}$ has entry $[\mathbf S_{n,m}]_{j,i}=s_{i,j,n}(f_m)$. The complex Tx and Rx response matrices, $\mathbf A_{n,m}^{\mathrm{Tx}}$ and $\mathbf A_{n,m}^{\mathrm{Rx}}$, have sizes $N_T\times N_v$ and $N_R\times N_v$, respectively. Each entry gives the one-way spherical-wave phase from an antenna element to a voxel:
\begin{equation}
\resizebox{\dimexpr\columnwidth-3em\relax}{!}{$\displaystyle
[\mathbf A_{n,m}^{\mathrm{Tx}}]_{i,g}=\ee^{-\iu\kappa_m r_{i,n}^{\mathrm{Tx}}(\mathbf q_g)},\quad
[\mathbf A_{n,m}^{\mathrm{Rx}}]_{j,g}=\ee^{-\iu\kappa_m r_{j,n}^{\mathrm{Rx}}(\mathbf q_g)}.
$}
\label{eq:spherical_array_matrices}
\end{equation}
The complete Tx--Rx observation has the matrix form
\begin{equation}
\mathbf S_{n,m}
=\mathbf A_{n,m}^{\mathrm{Rx}}\diag(\boldsymbol\rho)
(\mathbf A_{n,m}^{\mathrm{Tx}})^\T+\mathbf E_{n,m}.
\label{eq:mimo_matrix_model}
\end{equation}
The residuals form $\mathbf E_{n,m}$, using the same channel ordering as the measurement matrix. The transpose is unconjugated because the two propagation phases add along the path.

The operator $\vect(\cdot)$ stacks matrix columns with the Rx index varying fastest. Vectorizing $\mathbf S_{n,m}$ and $\mathbf E_{n,m}$ gives $\mathbf s_{n,m}$ and $\boldsymbol\epsilon_{n,m}$, respectively. The resulting complex sensing matrix $\boldsymbol\Phi_{n,m}$ maps $N_v$ voxel reflectivities to $N_TN_R$ measurements:
\begin{equation}
\mathbf s_{n,m}=\boldsymbol\Phi_{n,m}\boldsymbol\rho
+\boldsymbol\epsilon_{n,m}.
\label{eq:single_block_model}
\end{equation}
The $g$-th sensing-matrix column is given by
\begin{equation}
[\boldsymbol\Phi_{n,m}]_{:,g}
=\mathbf a_{n,m}^{\mathrm{Tx}}(\mathbf q_g)
\otimes\mathbf a_{n,m}^{\mathrm{Rx}}(\mathbf q_g),
\label{eq:nearfield_matrix_entry}
\end{equation}
where $\mathbf a_{n,m}^{\mathrm{Tx}}(\mathbf q_g)$ and $\mathbf a_{n,m}^{\mathrm{Rx}}(\mathbf q_g)$ are the $g$-th columns of the two response matrices, and $\otimes$ is the Kronecker product. Stack the blocks first over frequency and then over the $N_s$ frames. This gives the joint observation model
\begin{equation}
\mathbf s=\boldsymbol\Phi\boldsymbol\rho+\boldsymbol\epsilon,\qquad
\boldsymbol\Phi\in\mathbb C^{N_sN_fN_TN_R\times N_v}.
\label{eq:stacked_nearfield_model}
\end{equation}
\subsection{Wavefront Curvature and Geometric Limits}
\label{sec:observability}
A single antenna path illustrates the need for spherical propagation. Let $r_0>0$ denote the target range and $\mathbf u_0$ its unit direction from a reference position. Displacing the antenna by $\mathbf d$ changes the range to $r(\mathbf d)=\|r_0\mathbf u_0-\mathbf d\|_2$. For $\|\mathbf d\|_2\ll r_0$, a second-order expansion gives
\begin{equation}
r(\mathbf d)\simeq r_0-\mathbf u_0^\T\mathbf d
+\frac{\|\mathbf d\|_2^2-(\mathbf u_0^\T\mathbf d)^2}{2r_0}.
\label{eq:near_field_expansion}
\end{equation}
A plane-wave approximation retains the first two terms, omitting range-dependent curvature. In motion-assisted imaging, displacement includes physical antenna offsets and robot motion. The exact distances in~\eqref{eq:bistatic_range} avoid this linearization across the coherent aperture. Unlike angle-only far-field steering, spherical-wave focusing depends on both direction and distance. It remains limited by viewing diversity: forward motion supplies a predominantly horizontal synthetic aperture, while elevation information depends mainly on the physical array. These unequal resolutions motivate using reconstructed horizontal locations to guide height estimation.

\section{Bilateral 3D Reconstruction}
\label{sec:method}
Bilateral Tx combining precedes near-field back-projection (BP) and power fusion; multi-layer extraction and guarded MVDR then refine the height structure (Fig.~\ref{fig:diagram}).
Phase responses are computed without storing sensing matrices.

\subsection{Bilateral Tx Combining}
\label{sec:bilateral}
\subsubsection{Wideband directional steering}
Tx combining operates on calibrated MIMO data after acquisition. Center-frequency weights introduce phase mismatch at other frequencies, so steering weights are computed per frequency. This compensates wideband phase variation, not near-field wavefront curvature. For azimuth $\theta$ and elevation $\phi$, define the radar-frame unit direction vector
\begin{equation}
\mathbf u(\theta,\phi)
=[\sin\theta\cos\phi,\ \sin\phi,\ \cos\theta\cos\phi]^\T.
\label{eq:tx_direction_vector}
\end{equation}
The sign label $\sigma$ distinguishes the negative and positive lateral sides. Their azimuths are $\theta_-=-\theta_b$ and $\theta_+=\theta_b$, respectively. At elevation $\phi_b$, the direction for each side is $\mathbf u_\sigma=\mathbf u(\theta_\sigma,\phi_b)$. The complex vector $\mathbf w_m^\sigma$ contains one weight per Tx element:
\begin{equation}
[\mathbf w_m^\sigma]_i
=h_i\,\ee^{-\iu\kappa_m(\boldsymbol\delta_i^{\mathrm{Tx}})^\T\mathbf u_\sigma}.
\label{eq:wideband_tx_weight}
\end{equation}
The coefficients $h_i$ are nonnegative Hann weights normalized to sum to one. We use $\phi_b=0^\circ$ and $\theta_b=45^\circ$. These plane-wave steering weights depend on direction and frequency, not target range: they emphasize lateral echoes rather than focus at a near-field point. The operation is digital channel combining, not simultaneous hardware beam steering.

\subsubsection{Tx-channel combination}
The matrix $\mathbf W_m$ has $N_T$ rows and two columns: $\mathbf w_m^-$ for the negative side and $\mathbf w_m^+$ for the positive side. The bilateral measurements are
\begin{equation}
\widetilde{\mathbf S}_{n,m}
=\mathbf S_{n,m}\mathbf W_m
=[\widetilde{\mathbf s}_{n,m}^-,\widetilde{\mathbf s}_{n,m}^+].
\label{eq:bilateral_tx_combination}
\end{equation}
Each synthesized response retains all $N_R$ receivers. For side $\sigma$, its $j$-th sample is denoted $\widetilde s_{j,n}^\sigma(f_m)$. The original sensing model gives the same synthesized measurements
\begin{equation}
\widetilde{\mathbf s}_{n,m}^\sigma
=\big[(\mathbf w_m^\sigma)^\T\otimes\mathbf I_{N_R}\big]
\boldsymbol\Phi_{n,m}\boldsymbol\rho
+\widetilde{\boldsymbol\epsilon}_{n,m}^\sigma,
\label{eq:compressed_forward_model}
\end{equation}
where $\mathbf I_{N_R}$ is the Rx identity matrix and $\widetilde{\boldsymbol\epsilon}_{n,m}^\sigma$ denotes the combined residual. Use the negative-side response for $x<0$ and the positive-side response for $x\geq0$. Both responses preserve the Tx-path dependence encoded in $\mathbf W_m$.

\subsection{Near-Field Imaging and Fusion}
\label{sec:bp}
\subsubsection{Local apertures and paths}
For coherent integration, process the trajectory in overlapping windows
\begin{equation}
\mathcal W_\ell=\{n:\zeta_\ell\leq z_n\leq\zeta_\ell+L_a\},
\qquad \zeta_\ell=\zeta_0+\ell\Delta_a,
\label{eq:subaperture}
\end{equation}
where $L_a$ and $\Delta_a$ denote the window length and displacement, respectively, and $\zeta_0$ is the first window's starting position. The mean Tx center is the effective transmitter, with target range $r_n^0(\mathbf q)=\|\mathbf q-\mathbf p_n\|_2$. The effective bistatic path is
\begin{equation}
L_{j,n}(\mathbf q)=r_n^0(\mathbf q)+r_{j,n}^{\mathrm{Rx}}(\mathbf q).
\label{eq:effective_bistatic_path}
\end{equation}
The exact Tx-compressed kernel factors into the effective-path phase $\ee^{-\iu\kappa_mL_{j,n}(\mathbf q)}$ and the residual Tx factor
\begin{equation}
\gamma_{n,m}^\sigma(\mathbf q)=
\sum_{i=1}^{N_T}[\mathbf w_m^\sigma]_i
\ee^{-\iu\kappa_m[r_{i,n}^{\mathrm{Tx}}(\mathbf q)-r_n^0(\mathbf q)]}.
\label{eq:tx_pattern_factor}
\end{equation}
BP compensates the candidate-dependent phase of $L_{j,n}$ but leaves the residual Tx factor $\gamma_{n,m}^\sigma$ uncompensated. It therefore provides near-field focusing under an effective-Tx approximation, rather than exact matching of the full Tx-compressed response. Height refinement instead uses original channels and exact bistatic distances.

\subsubsection{Range compression and focusing}
Range compression uses a Hann frequency window $h_f[m]$, then an $N_b$-point inverse discrete Fourier transform (IDFT). With $N_b\geq N_f$ allowing zero-padding, the synthesized range samples are
\begin{equation}
b_{j,n}^\sigma[b]=\frac{1}{N_b}\sum_{m=0}^{N_f-1}
h_f[m]\widetilde s_{j,n}^\sigma(f_m)\ee^{\iu2\pi mb/N_b},
\label{eq:range_profile}
\end{equation}
for $b=0,\ldots,N_b-1$. Their one-way range coordinates are $r_b=cb/(2N_b\Delta f)$. Denote piecewise-linear interpolation of these complex samples by $\bar b_{j,n}^\sigma(r)$. Zero-padding reduces sample spacing but does not increase physical range resolution. The voxel-aligned complex sample is
\begin{equation}
v_{j,n}^\sigma(\mathbf q)
=\bar b_{j,n}^\sigma\!\left(\frac{L_{j,n}(\mathbf q)}{2}\right)
\ee^{\iu\kappa_0L_{j,n}(\mathbf q)}.
\label{eq:bp_aligned_sample}
\end{equation}
Sampling at half the total path aligns the wideband delay, while the exponential compensates the residual phase at $f_0$. At voxel $\mathbf q$, select the lateral response using $\sigma(\mathbf q)$. Stack its valid samples from window $\mathcal W_\ell$ into $\mathbf v_\ell(\mathbf q)$, with the Rx index varying fastest. The corresponding nonnegative Rx and motion Hann weights form $\boldsymbol\omega_\ell(\mathbf q)$, normalized to sum to one. The inner product gives the local image
\begin{equation}
B_\ell(\mathbf q)=
\boldsymbol\omega_\ell^\T(\mathbf q)\mathbf v_\ell(\mathbf q).
\label{eq:local_nearfield_bp}
\end{equation}
Both vectors index valid receiver--frame samples, with weights normalized after excluding out-of-range samples. Windows without valid samples are omitted from fusion.

\subsubsection{Top-\texorpdfstring{$K$}{K} power fusion}
Define the local power as $P_\ell(\mathbf q)=|B_\ell(\mathbf q)|^2$. With changing target visibility, averaging all windows can dilute well-focused responses, while a single maximum can emphasize isolated artifacts. Let $\mathcal K(\mathbf q)$ index up to $K$ windows with the strongest valid powers, and write $K_{\mathbf q}=|\mathcal K(\mathbf q)|$. For $K_{\mathbf q}>0$, fuse them as
\begin{equation}
I(\mathbf q)=\frac{1}{K_{\mathbf q}}
\sum_{\ell\in\mathcal K(\mathbf q)}P_\ell(\mathbf q).
\label{eq:topk_spatial_fusion}
\end{equation}
Set $I(\mathbf q)=0$ when no window has valid support. Fusion is coherent within each local aperture and incoherent across apertures. Overlapping windows reuse measurements, so fused power is an imaging response, not a statistical confidence score. The volume supports detection and height extraction.

\subsection{Multi-Layer Reconstruction}
\label{sec:layers}
\subsubsection{Lateral--forward support}
Sample the reconstruction grid at lateral coordinates $x_u$, heights $y_v$, and forward positions $z_w$, using $N_x$, $N_y$, and $N_z$ samples along the respective axes. Store the nonnegative voxel intensities in the tensor $\mathcal I$ of size $N_x\times N_y\times N_z$. Its entry $\mathcal I_{u,v,w}$ equals $I([x_u,y_v,z_w]^\T)$. Maximizing over height gives the projection matrix $\mathbf M$, with $N_x$ rows and $N_z$ columns:
\begin{equation}
[\mathbf M]_{u,w}=\max_{1\leq v\leq N_y}\mathcal I_{u,v,w}.
\label{eq:intensity_tensor}
\end{equation}
A two-dimensional constant-false-alarm-rate (CFAR) detector identifies high-confidence seeds in $\mathbf M$. Bounded neighborhood expansion with a local peak-relative threshold recovers the surrounding spatial extent. The set $\mathcal H$ contains retained grid locations $(u_k,w_k)$, indexed by $k=1,\ldots,N_h$. Detection in this projection uses the more stable lateral--forward response before resolving its vertical components.

\subsubsection{Height profiles and layer extraction}
The height-profile matrix $\mathbf P_y$ has $N_y$ height samples and $N_h$ support locations:
\begin{equation}
[\mathbf P_y]_{v,k}=\mathcal I_{u_k,v,w_k}.
\label{eq:vertical_profile}
\end{equation}
Column $k$ contains the height profile at $(x_{u_k},z_{w_k})$. Its strongest peak provides the primary height candidate. Additional local maxima must exceed a threshold relative to the dominant peak, have sufficient prominence over neighboring troughs, and maintain a minimum height separation from already retained peaks. This allows one lateral--forward location to contain multiple scattering layers rather than forcing it to have one height. Secondary candidates are also checked across neighboring support locations. At adjacent locations, candidate heights $y_{k,r}$ and $y_{k',r'}$ are consistent when their absolute difference does not exceed $\delta_y$. Only secondary components with sufficient spatial persistence and separation from the height-search boundaries are kept. Collecting the primary and accepted secondary components gives $N_p$ BP points. For point $p$, the estimate $\widehat{\mathbf q}_p^{\mathrm{BP}}$ has lateral coordinate $x_p$, height $y_p^{\mathrm{BP}}$, and forward coordinate $z_p$.

\subsection{Guarded Near-Field MVDR Refinement}
\label{sec:mvdr}
\subsubsection{Local search and sampling}
For point $p$, keep $x_p$ and $z_p$ fixed and evaluate $\mathbf q_p(y)=[x_p,y,z_p]^\T$ on the height grid
\begin{equation}
\mathcal Y_p=\{y_p^{\mathrm{BP}}+t\delta_{\mathrm{MV}}:
t\in\mathbb Z,\ |t\delta_{\mathrm{MV}}|\leq\Delta_y\}.
\label{eq:local_height_grid}
\end{equation}
The search half-span is $\Delta_y$, and the height step is $\delta_{\mathrm{MV}}$. Refinement uses all original $N_TN_R$ channels before Tx combining. Apply IDFT and interpolation to each channel to obtain its complex range profile $\bar b_{i,j,n}(r)$. Snapshot index $s$ identifies a valid Tx--frame pair $(i_s,n_s)$ within the local aperture, with $s=1,\ldots,N_{\mathrm{snap}}$. Each pair defines an $N_R$-entry complex Rx snapshot $\mathbf c_{p,s}(y)$:
\begin{equation}
[\mathbf c_{p,s}(y)]_j
=\bar b_{i_s,j,n_s}\!\left(
\frac{R_{i_s,j,n_s}(\mathbf q_p(y))}{2}\right).
\label{eq:raw_mvdr_sample}
\end{equation}
Each trial height defines a spatial focal point. Unlike front-end BP, range sampling uses the separate physical Tx and Rx paths. $N_{\mathrm{snap}}$ counts valid Tx--frame snapshots at that height.

\subsubsection{Snapshot focusing}
Tx--frame propagation phases must be aligned before covariance estimation. Choose the central frame $n_0$ as the reference. The vectors $\mathbf d_{p,s}(y)$ and $\mathbf a_p(y)$ describe the original bistatic response and reference Rx response, respectively. Their $N_R$ complex entries are
\begin{equation}
\begin{aligned}
\relax[\mathbf d_{p,s}(y)]_j
&=\ee^{-\iu\kappa_0 R_{i_s,j,n_s}(\mathbf q_p(y))},\\
[\mathbf a_p(y)]_j
&=\ee^{-\iu\kappa_0 r_{j,n_0}^{\mathrm{Rx}}(\mathbf q_p(y))}.
\end{aligned}
\label{eq:nearfield_rx_manifold}
\end{equation}
The diagonal focusing matrix and focused snapshot are
\begin{equation}
\mathbf F_{p,s}(y)=
\diag\!\big(\mathbf a_p(y)\odot\mathbf d_{p,s}^*(y)\big),
\label{eq:mvdr_reference_focusing}
\end{equation}
\begin{equation}
\mathbf z_{p,s}(y)=\mathbf F_{p,s}(y)\mathbf c_{p,s}(y),
\label{eq:focused_snapshot}
\end{equation}
where $\odot$ denotes element-wise multiplication. Every entry of $\mathbf d_{p,s}$ has unit magnitude, so the focusing matrix maps $\mathbf d_{p,s}$ to $\mathbf a_p$. A matched scatterer with complex amplitude $\alpha_{p,s}$ therefore yields the focused response $\alpha_{p,s}\mathbf a_p$, apart from interference and noise. Tx and motion offsets are removed; reference-frame Rx phase differences are preserved.

Collect the aligned snapshots into
\begin{equation}
\mathbf Z_p(y)
=[\mathbf z_{p,1}(y),\ldots,\mathbf z_{p,N_{\mathrm{snap}}}(y)]
\in\mathbb C^{N_R\times N_{\mathrm{snap}}}.
\label{eq:mvdr_snapshot_matrix}
\end{equation}
Each column is a focused Tx--frame snapshot. The $N_R\times N_R$ covariance acts on receiver coordinates; transmitters add snapshots, not adaptive-array dimensions or vertical aperture.

\begin{figure}[t]
\centering
\begin{subfigure}[b]{0.495\linewidth}
\centering
\includegraphics[width=\linewidth,height=0.95\linewidth,keepaspectratio]{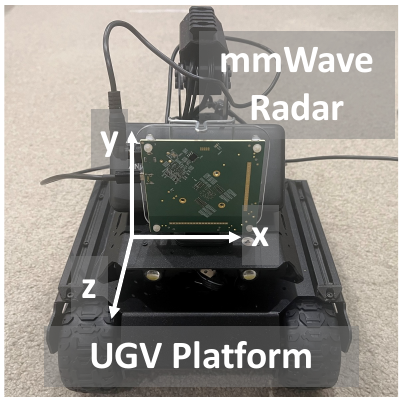}
\caption{UGV platform.}
\label{fig:platform-ugv}
\end{subfigure}\hfill
\begin{subfigure}[b]{0.495\linewidth}
\centering
\includegraphics[width=\linewidth,height=0.95\linewidth,keepaspectratio]{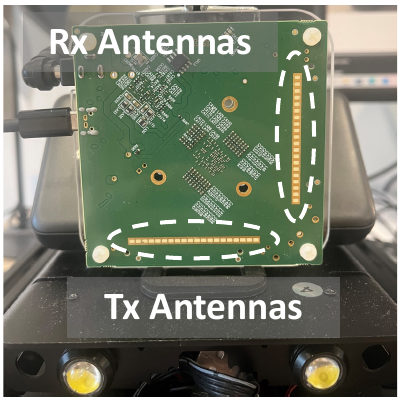}
\caption{MIMO radar.}
\label{fig:platform-radar}
\end{subfigure}
\caption{UGV platform and MIMO radar.}
\label{fig:platform}
\vspace{-1.5em}
\end{figure}

\begin{figure*}[t]
\centering
\begin{minipage}[t]{0.185\linewidth}
\centering
\includegraphics[width=\linewidth]{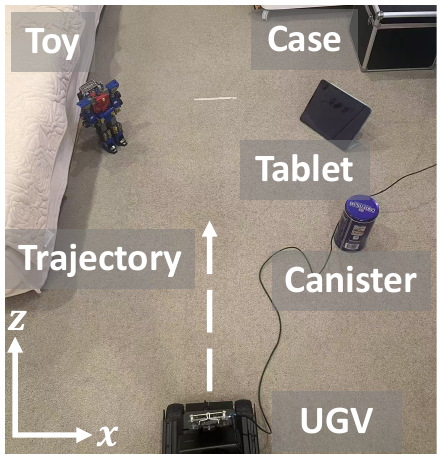}
\subcaption{Controlled test scene.}
\label{fig:controlled_scene}
\end{minipage}
\begin{minipage}[t]{0.27\linewidth}
\centering
\includegraphics[width=\linewidth]{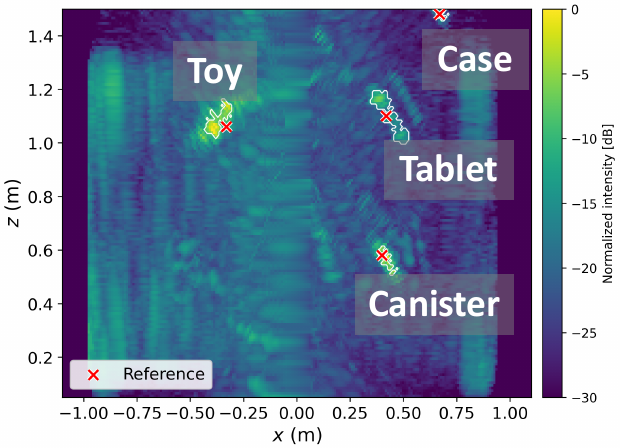}
\subcaption{Radar intensity map.}
\label{fig:controlled_xz}
\end{minipage}
\begin{minipage}[t]{0.255\linewidth}
\centering
\includegraphics[width=\linewidth]{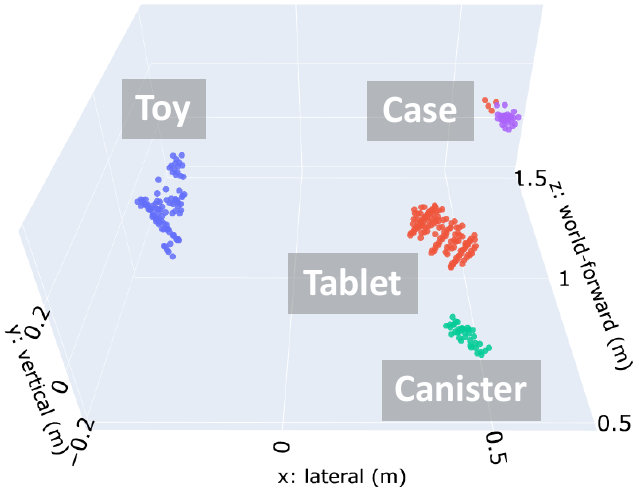}
\subcaption{3D reconstruction.}
\label{fig:controlled_3d}
\end{minipage}
\begin{minipage}[t]{0.27\linewidth}
\centering
\includegraphics[width=\linewidth]{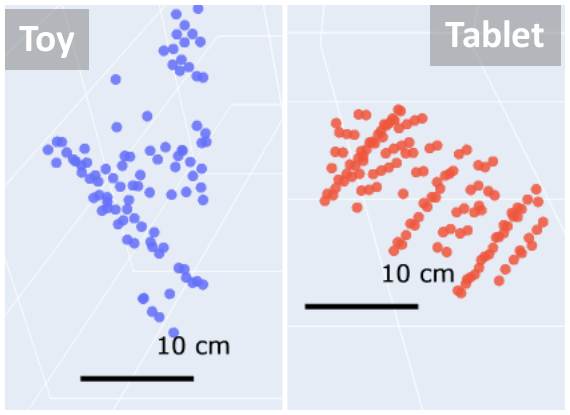}
\subcaption{Toy and tablet close-ups.}
\label{fig:controlled_zoom}
\end{minipage}
\caption{Controlled-scene results: red crosses mark measured references; point colors distinguish targets.}
\label{fig:controlled_validation}
\vspace{-.5em}
\end{figure*}

\begin{figure*}[t]
\centering
\begin{minipage}[t]{0.495\linewidth}
\centering
\includegraphics[width=0.85\linewidth]{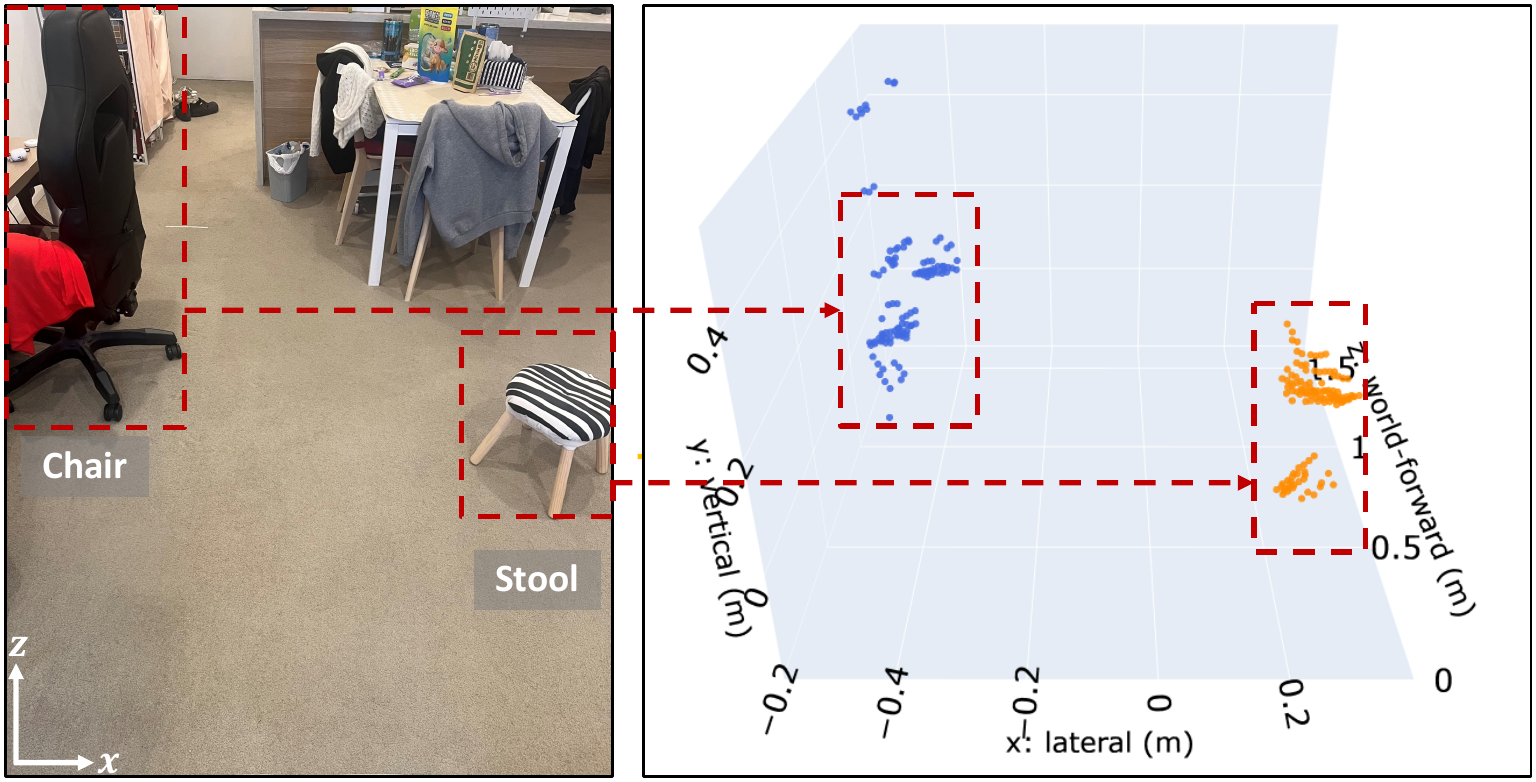}
\subcaption{Office/study scene.}
\label{fig:office_reconstruction}
\end{minipage}
\begin{minipage}[t]{0.495\linewidth}
\centering
\includegraphics[width=\linewidth]{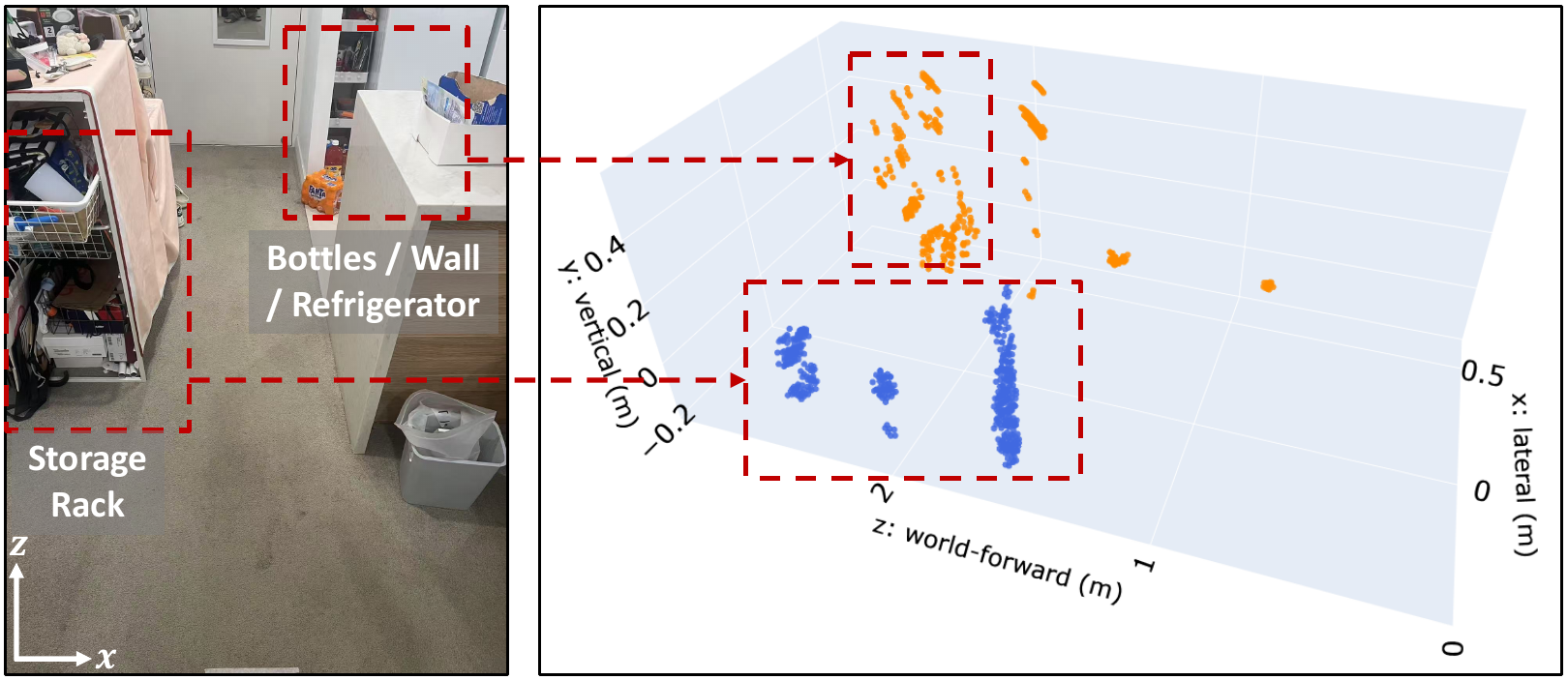}
\subcaption{Corridor scene.}
\label{fig:corridor_reconstruction}
\end{minipage}
\caption{Indoor 3D reconstruction: blue and orange points correspond to $x<0$ and $x\geq0$, respectively.}
\label{fig:indoor_reconstruction}
\vspace{-1em}
\end{figure*}

\subsubsection{Covariance and MVDR spectrum}
The sample covariance of the focused snapshots is
\begin{equation}
\widehat{\mathbf R}_p(y)
=\frac{1}{N_{\mathrm{snap}}}\mathbf Z_p(y)\mathbf Z_p^\HH(y).
\label{eq:nearfield_covariance}
\end{equation}
Trace-normalized diagonal loading gives $\overline{\mathbf R}_p(y)=\widehat{\mathbf R}_p(y)+\lambda_p(y)\mathbf I_{N_R}$, where $\lambda_p(y)=\alpha_{\mathrm{DL}}\tr(\widehat{\mathbf R}_p(y))/N_R$, $\alpha_{\mathrm{DL}}>0$, and $\tr(\cdot)$ denotes the trace. After excluding heights with zero energy or insufficient snapshots, we evaluate
\begin{equation}
P_p^{\mathrm{MVDR}}(y)
=\frac{1}{\mathbf a_p^\HH(y)\overline{\mathbf R}_p^{-1}(y)\mathbf a_p(y)}.
\label{eq:nearfield_mvdr}
\end{equation}
We evaluate the denominator by solving a linear system. Range sampling and focusing depend on $y$, so the covariance is recomputed at each trial height. The spectrum is used only to score local BP height candidates.

\subsubsection{Guarded peak selection}
Reliable local maxima of $P_p^{\mathrm{MVDR}}$ form the set $\mathcal P_p$ and must lie strictly inside the search interval $\mathcal Y_p$. When the snapshot and covariance checks pass, select the peak nearest to $y_p^{\mathrm{BP}}$, resolving equal-distance peaks by their spectral value. If no peak passes these checks, retain the original BP height:
\begin{equation}
\widehat y_p=
\begin{cases}
\displaystyle\argmin_{y\in\mathcal P_p}|y-y_p^{\mathrm{BP}}|,
&\text{valid refinement},\\[1mm]
y_p^{\mathrm{BP}},&\text{otherwise}.
\end{cases}
\label{eq:guarded_height_update}
\end{equation}
The final point is $\widehat{\mathbf q}_p=[x_p,\widehat y_p,z_p]^\T$. The update retains each point's index and lateral--forward coordinates, preserves the point count, and bounds its height displacement by $\Delta_y$. Independent height updates impose no separation constraint.

\section{Experiments}
We evaluate localization and background suppression in controlled household-object and indoor scenes, and qualitatively examine the effects of height refinement.

\subsection{Experimental Setup}
\subsubsection{Platform and parameters}
The UGV in Fig.~\ref{fig:platform-ugv} carries a Vayyar IMAGEVK-74 SFCW radar equipped with 20 Tx and 20 Rx elements (Fig.~\ref{fig:platform-radar}). The measurements use a fixed subset of 10 Tx elements and all 20 Rx elements; hence, $N_T=10$ and $N_R=20$ throughout the formulation and processing.
Coordinates follow Section~\ref{sec:geometry}: lateral $x$, vertical $y$, and forward $z$, aligned with the radar boresight and motion. Each approximately $30$-ms sweep records $N_f=41$ uniformly spaced frequencies spanning $6$~GHz, corresponding to a nominal range resolution of 2.5 cm; $\Delta f=150$~MHz and a single-channel range ambiguity period of $c/(2\Delta f)\approx1$~m. Each frame contributes only within approximately $1$~m of its radar position; combining local observations along the trajectory extends the world-frame coverage. The UGV follows an approximately straight path, with nominal frame positions determined from its known speed and motion during each sweep treated as negligible. We evaluate three scenes: four controlled targets, an office/study, and a longer corridor. Object or landmark reference positions are measured manually. Unless stated otherwise, the coherent aperture length is $L_a=0.40$~m, the aperture displacement is $0.10$~m, and the 3D reconstruction grid spacing is $1$~cm. Bilateral Tx directions are $\theta_\sigma=\pm45^\circ$ with $\phi_b=0^\circ$, and fusion uses $K=3$. 
Lateral--forward support uses 2-D CA-CFAR with $P_{\rm FA}=10^{-4}$, 3-cm guard and 7-cm training regions, followed by region growing at $-8$ dB relative to the local peak. Secondary peaks must lie within $-8$ dB of the primary, be separated by at least 5 cm with 1.5-dB prominence, and belong to a consistent cluster spanning at least three $x$--$z$ support cells within a 2.5-cm height tolerance.
Guarded MVDR refinement uses a search half-span $\Delta_y=5$~cm, a height step $\delta_{\mathrm{MV}}=5$~mm, and a loading factor $\alpha_{\mathrm{DL}}=0.10$. BP and MVDR use the same support and initial heights.

\begin{figure*}[t]
\centering
\begin{minipage}[t]{0.325\linewidth}
\centering
\includegraphics[width=\linewidth]{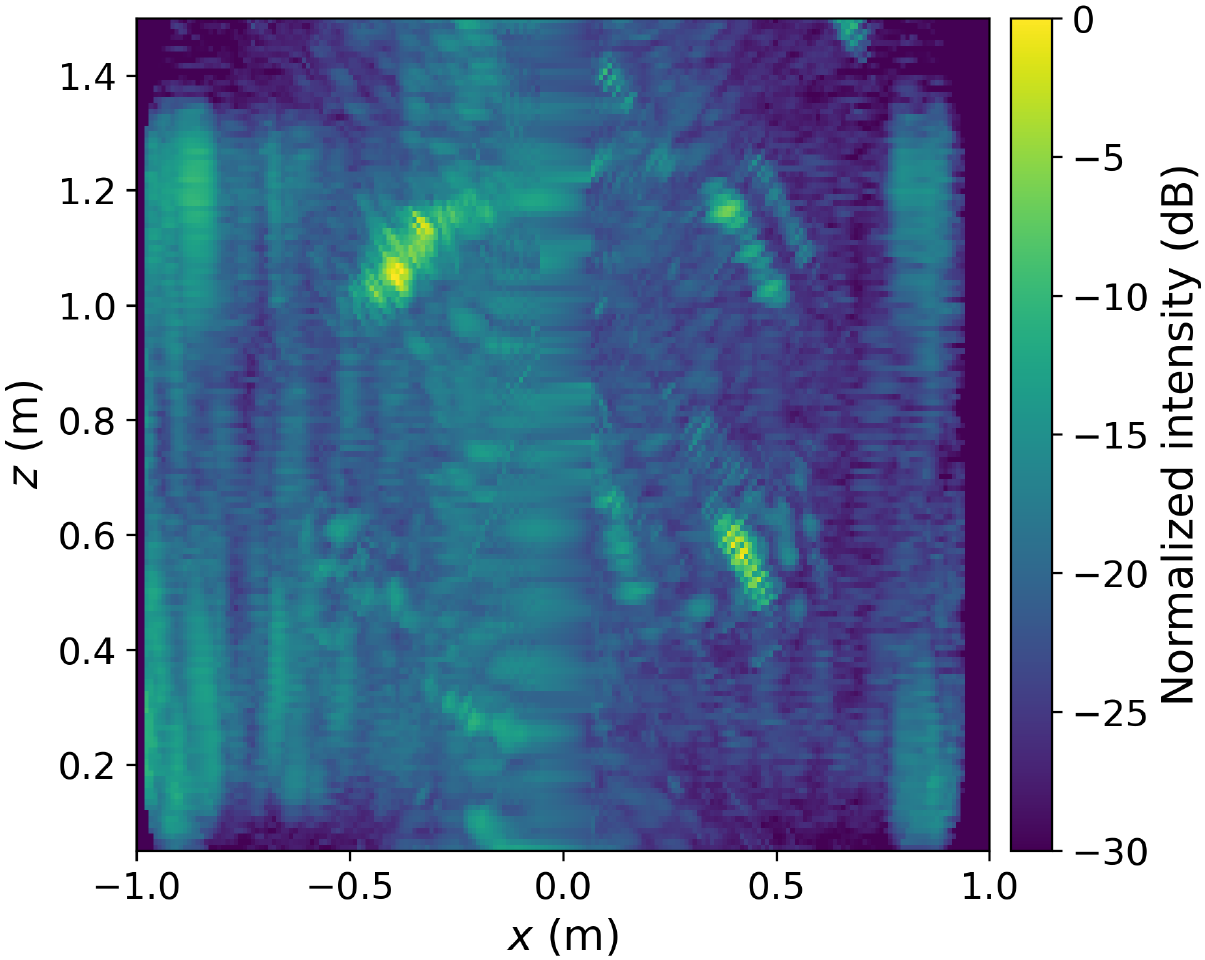}
\subcaption{BilatSAR.}
\label{fig:proposed-xz}
\end{minipage}
\begin{minipage}[t]{0.325\linewidth}
\centering
\includegraphics[width=\linewidth]{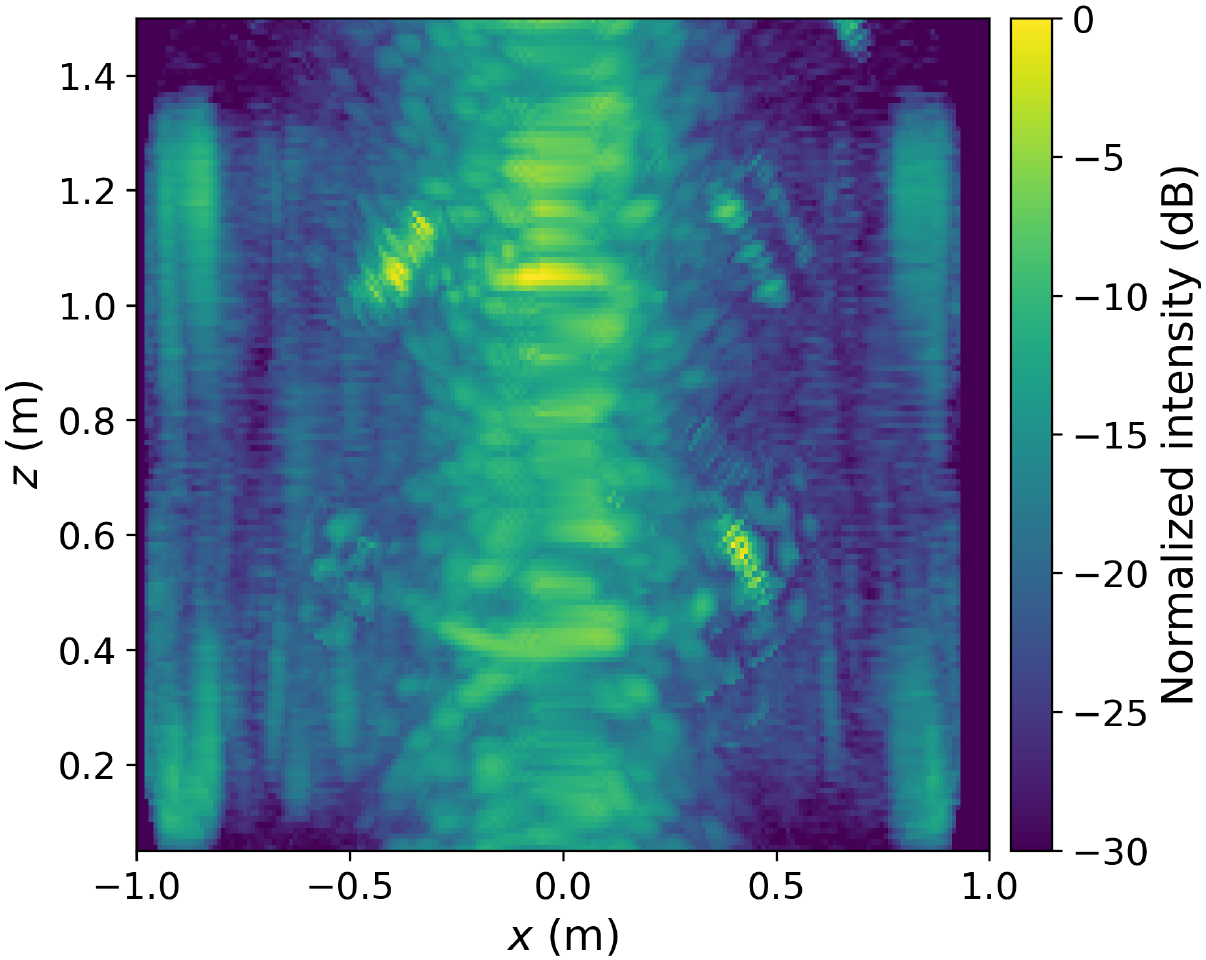}
\subcaption{No bilateral Tx combining.}
\label{fig:no-bf-xz}
\end{minipage}
\begin{minipage}[t]{0.325\linewidth}
\centering
\includegraphics[width=\linewidth]{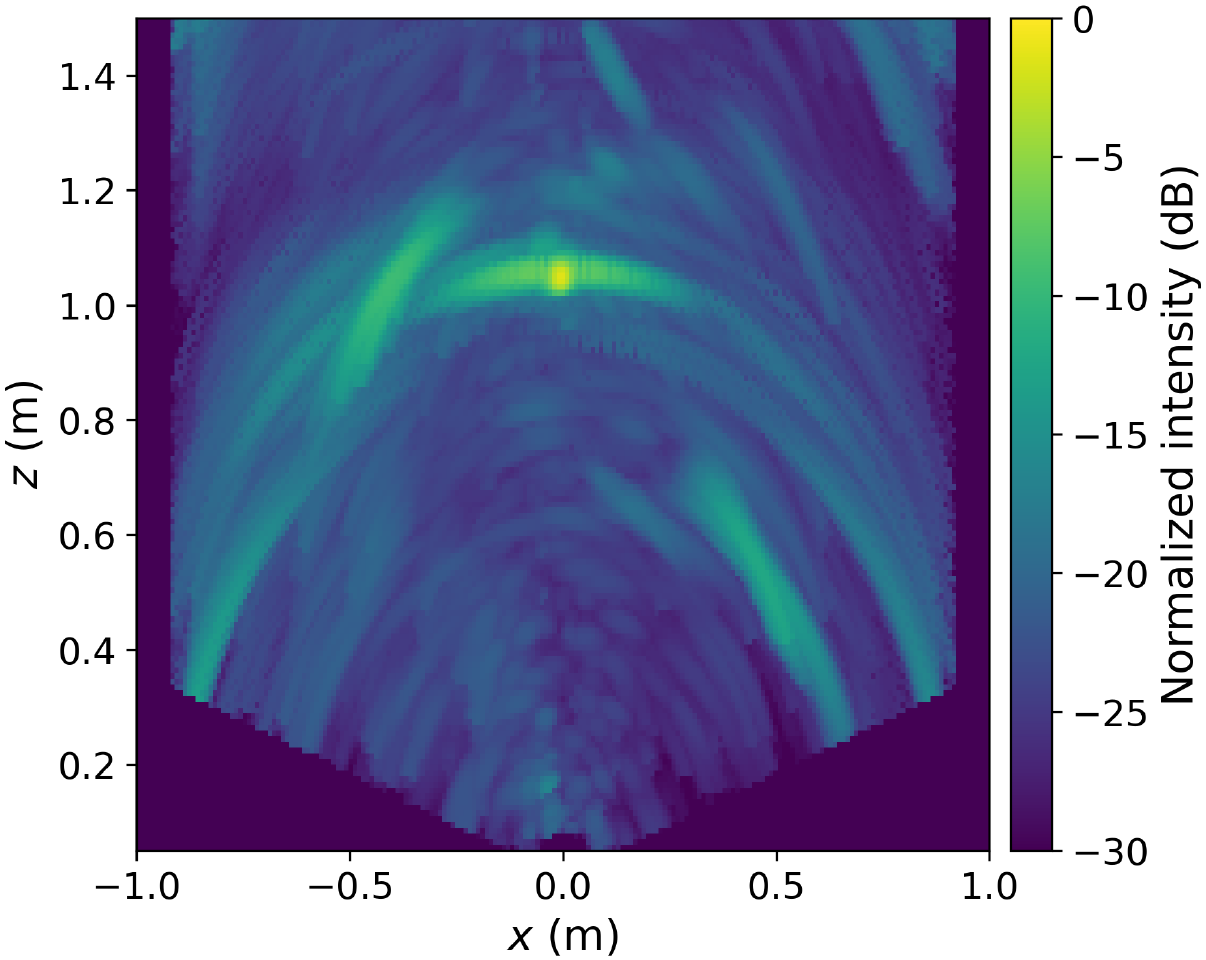}
\subcaption{MIMO Range--AoA.}
\label{fig:mimo-xz}
\end{minipage}
\caption{Bilateral Tx ablation and frame-wise MIMO Range--AoA comparison.}
\label{fig:frontend_comparison}
\vspace{-.5em}
\end{figure*}

\begin{figure*}[t]
\centering
\begin{minipage}[t]{0.325\linewidth}
\centering
\includegraphics[width=\linewidth]{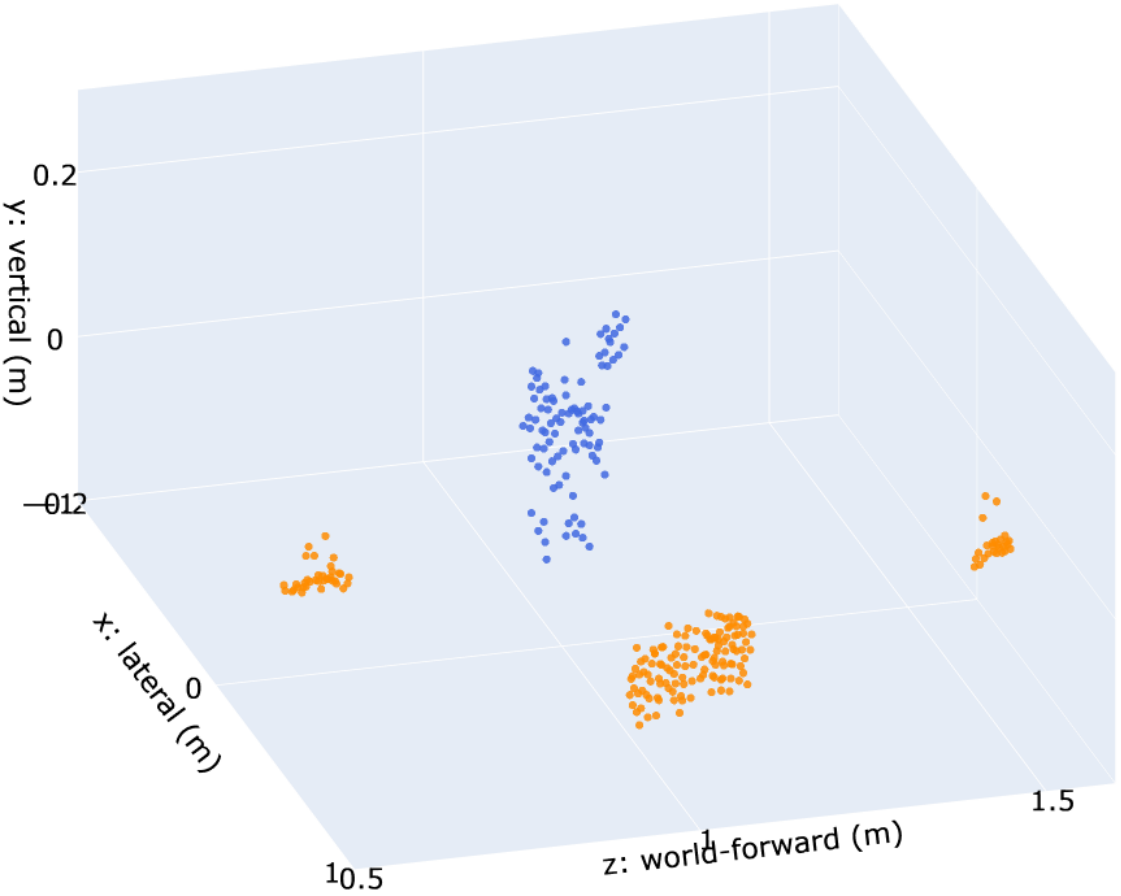}
\subcaption{Guarded MVDR refinement.}
\label{fig:3d-mvdr}
\end{minipage}
\begin{minipage}[t]{0.325\linewidth}
\centering
\includegraphics[width=\linewidth]{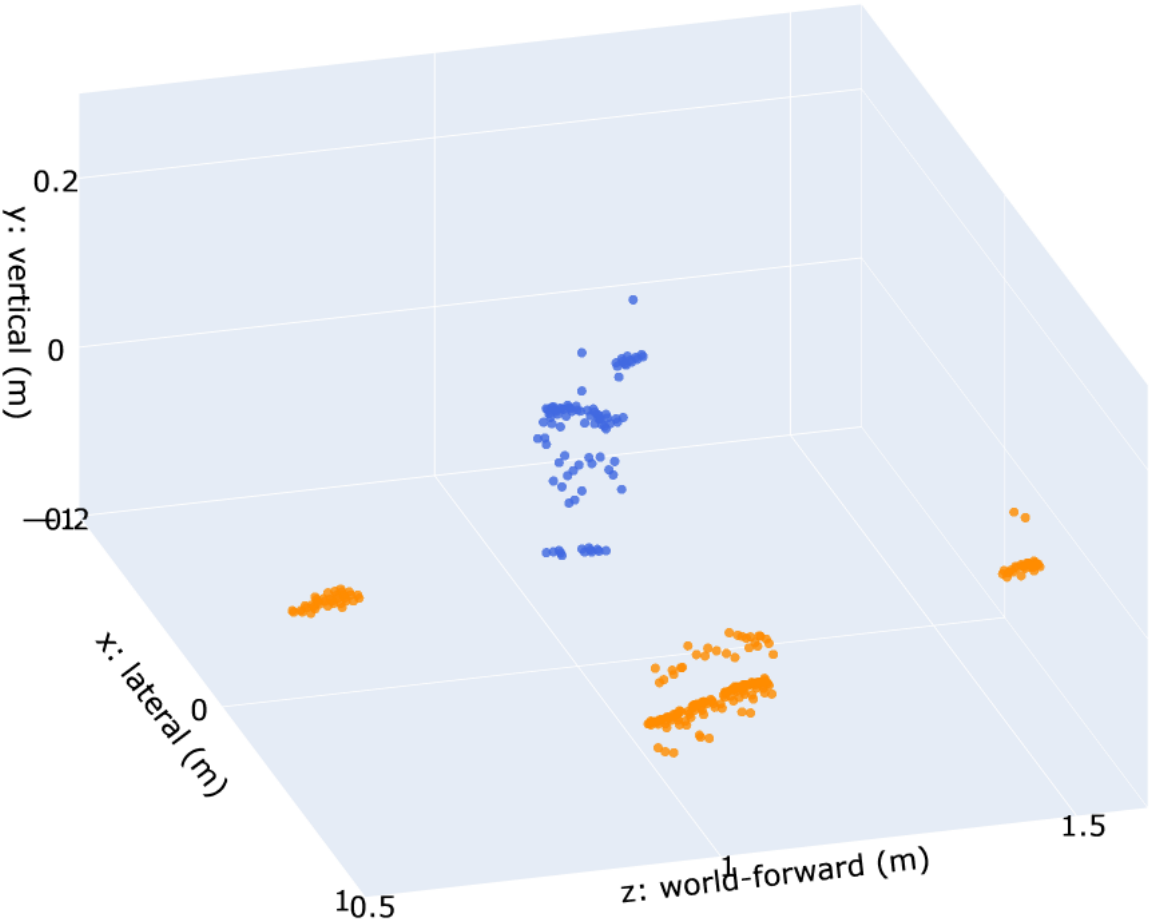}
\subcaption{Multi-layer BP reconstruction.}
\label{fig:3d-bp}
\end{minipage}
\begin{minipage}[t]{0.325\linewidth}
\centering
\includegraphics[width=\linewidth]{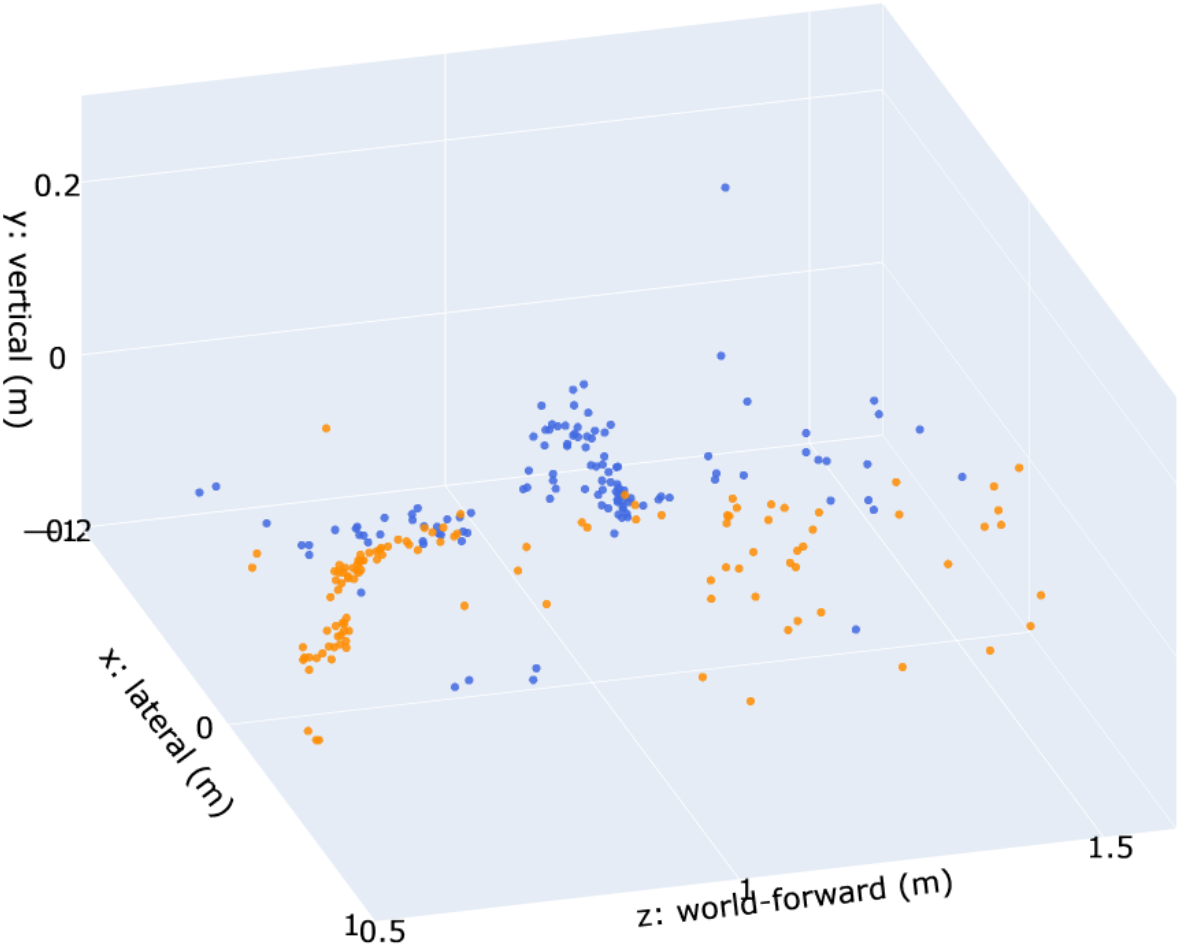}
\subcaption{MIMO with 3D CFAR.}
\label{fig:3d-mimo}
\end{minipage}
\caption{3D reconstruction with guarded MVDR, multi-layer BP, and conventional MIMO.}
\label{fig:3d_comparison}
\vspace{-1.em}
\end{figure*}

\begin{table}[t]
\centering
\caption{Lateral--forward errors and PBR.}
\label{tab:quantitative_comparison}
\setlength{\tabcolsep}{5.4pt}
\renewcommand{\arraystretch}{1.00}
\footnotesize
\begin{tabular}{llccc}
\toprule
Scene & Method & $\overline e_{xz}$ [cm] & $e_{xz}^{\max}$ [cm] & PBR [dB]\\
\midrule
\multirow{3}{*}{Controlled}
& BilatSAR & 4.65 & 7.62 & 15.59\\
& w/o Tx combining & 3.51 & 7.62 & 11.50\\
& Range--AoA & 8.55 & 14.21 & 6.43\\
\midrule
\multirow{3}{*}{Office/study}
& BilatSAR & 5.63 & 7.42 & 13.97\\
& w/o Tx combining & 5.96 & 7.91 & 10.66\\
& Range--AoA & 10.74 & 20.34 & 4.42\\
\midrule
\multirow{3}{*}{Corridor}
& BilatSAR & 5.98 & 6.95 & 13.02\\
& w/o Tx combining & 7.74 & 15.64 & 9.16\\
& Range--AoA & 20.23 & 45.54 & 4.38\\
\bottomrule
\end{tabular}
\vspace{-1em}
\end{table}

\subsubsection{Baselines and metrics}
The variant without bilateral Tx combining omits this stage but retains motion-assisted reconstruction. The conventional baseline performs frame-wise MIMO range--angle-of-arrival (Range--AoA) processing and maps the resulting detections into the world frame. For visualization, intensity maps cover the same spatial region and use a 30 dB dynamic range, each normalized to its own peak. The ablation isolates bilateral Tx combining; the Range--AoA comparison changes both aperture integration and angular processing, so it cannot isolate spherical-wave modeling. For each matched target or landmark $t$, let $\widehat{\mathbf u}_t=[\widehat x_t,\widehat z_t]^\T$ and $\mathbf u_t^{\mathrm{ref}}=[x_t^{\mathrm{ref}},z_t^{\mathrm{ref}}]^\T$ be the reconstructed and measured lateral--forward positions. We evaluate the $N_{\mathrm{ref}}$ matched references using lateral--forward errors:
\begin{equation}
\resizebox{\dimexpr\columnwidth-3em\relax}{!}{$\displaystyle
e_t=\|\widehat{\mathbf u}_t-\mathbf u_t^{\mathrm{ref}}\|_2,\quad
\overline e_{xz}=\frac{1}{N_{\mathrm{ref}}}
\sum_{t=1}^{N_{\mathrm{ref}}}e_t,\quad
e_{xz}^{\max}=\max_{1\leq t\leq N_{\mathrm{ref}}}e_t.
$}
\label{eq:localization_metrics}
\end{equation}
PBR is $10\log_{10}(P_{\mathrm{pk}}/P_{\mathrm{bg}})$, where $P_{\mathrm{pk}}$ is the local target peak and $P_{\mathrm{bg}}$ is the mean linear background power outside target neighborhoods; Table I reports the mean target-wise PBR in each scene.
Height refinement is assessed qualitatively; height accuracy and surface completeness are not quantified.

\subsection{Controlled-Scene Validation}
The controlled scene in Fig.~\ref{fig:controlled_scene} contains four targets on both sides of the trajectory. Their approximate $x$--$z$ reference positions are $(-0.33,1.06)$~m for the humanoid toy, $(0.40,0.58)$~m for the canister, $(0.42,1.10)$~m for the tablet, and $(0.67,1.48)$~m for the black case. Toy and tablet references are measured near their geometric centers; larger containers use the centers of their radar-facing surfaces. A radar intensity maximum represents a dominant scattering response, which need not coincide with an object's geometric center. These references support spatial comparison, not exact scatterer localization. Fig. ~\ref{fig:controlled_xz} shows separated lateral--forward support for all four targets. Clusters remain distinguishable after height reconstruction (Fig.~\ref{fig:controlled_3d}), with vertically distributed echoes in the toy and tablet details (Fig.~\ref{fig:controlled_zoom}). In this pipeline, motion-assisted BP establishes spatial support, and the original MIMO measurements provide the local vertical evidence. The reconstructed points should therefore be interpreted as a scattering representation of the objects, not as uniformly sampled physical surfaces.

\subsection{Indoor Environment Sensing}
Fig.~\ref{fig:indoor_reconstruction} evaluates distributed structure in an office/study and a corridor containing furniture and clutter. In the office/study, clusters correspond to structures on both sides of the path, including the chair and stool. In the corridor, separated responses extend over a longer forward range, including the storage rack and the bottles/wall/refrigerator region. Nonuniform height distributions indicate that the reconstruction is not restricted to a prescribed plane. Local-aperture imaging and Top-$K$ fusion address varying visibility, preserving clusters corresponding to structures on both sides.

\subsection{Ablation and Baseline Comparison}
\subsubsection{Bilateral Tx combining}
Fig.~\ref{fig:frontend_comparison} compares BilatSAR, the variant without bilateral Tx combining, and frame-wise MIMO Range--AoA. BilatSAR produces compact, separated target responses with a relatively clean background. Removing bilateral Tx combining increases diffuse responses and sidelobe-like structures, including a pronounced response near the trajectory at $x=0$. This response is consistent with greater residual platform or self-response, although the figure does not identify its source independently. The conventional MIMO result exhibits broad arc-shaped smearing after detections are mapped into the world frame, making neighboring structures less distinguishable. Together, these results support bilateral Tx combining for background suppression and coherent motion integration for spatial concentration in the evaluated sensing configuration.

\subsubsection{Local height refinement}
Fig.~\ref{fig:3d_comparison} compares multi-layer BP with guarded near-field MVDR using the same detected support and candidate set. BP produces discrete vertical layers, whereas the adaptive stage locally adjusts candidate heights and yields a more continuous distribution. This stage neither changes lateral--forward coordinates nor introduces additional detections. The conventional baseline is extended to 3D by estimating elevation at its cell-averaging CFAR detections; several dominant target regions remain visible, but the cloud contains more scattered and isolated responses. This comparison shows local height adjustments at fixed lateral--forward support; it does not measure changes in vertical accuracy or resolution.

\subsubsection{Localization and background rejection}
Table~\ref{tab:quantitative_comparison} reports errors and PBR. BilatSAR gives the highest PBR in all three scenes and lower mean and maximum errors than Range--AoA throughout. The controlled scene is the only case in which the variant without bilateral Tx combining has a lower mean error ($3.51$ vs.\ $4.65$~cm), but its PBR is $4.09$~dB lower; thus, bilateral Tx combining mainly improves target-to-background separation there. In the office/study, BilatSAR improves both error and PBR over this variant. The methods differ more in the corridor, where bilateral Tx combining reduces the maximum error from $15.64$ to $6.95$~cm and raises PBR by $3.86$~dB, while Range--AoA exceeds $20$~cm mean error. The results support the combined use of bilateral Tx combining and coherent near-field imaging for localization and background rejection in the tested static scenes.

\section{Conclusions}
This paper presented BilatSAR, a physics-based framework for bilateral 3D robot sensing with one forward-mounted MIMO radar. A spherical-wave measurement model relates scene reflectivity to radar observations across antennas, frequencies, and robot positions, distinguishing exact bistatic measurements from the effective-Tx approximation used in the imaging front end. Bilateral Tx combining emphasizes lateral directions, while sliding-aperture near-field BP and Top-$K$ fusion establish lateral--forward support; multi-layer recovery extracts height candidates for local guarded MVDR refinement. Controlled, office/study, and corridor experiments show compact bilateral scattering structure, with mean lateral--forward errors below $6$~cm and PBR above $13$~dB. These results support near-field robot sensing without extra radars or learned scene priors. Future work will study trajectory uncertainty and dynamic scenes, using independent 3D ground truth to assess height.

\bibliographystyle{IEEEtran}
\bibliography{IEEEabrv,refs_v2_ZQ}
\end{document}